\documentclass[preprint,12pt]{elsarticle}

\usepackage{amssymb}
\usepackage{epsfig}
\usepackage{amsmath}

\usepackage{graphicx}       
\usepackage{multirow}       
\usepackage{float}
\numberwithin{equation}{section}

\usepackage{lineno}
\usepackage{caption}
\usepackage{float} 
\usepackage{subcaption}
\journal{}
\usepackage{hyperref}
\hypersetup{
	colorlinks = true, 
	allcolors = blue,  
}
\usepackage[nameinlink]{cleveref}

\crefname{figure}{Fig.}{Figs.}
\crefformat{equation}{Eq.~#2(#1)#3}
\crefformat{section}{Section~#2#1#3}
\AtBeginDocument{%
	\let\citet\cite
}

\usepackage[labelfont=bf]{caption}
\begin{document}
\biboptions{sort&compress}
\date{}	
\begin{frontmatter}
\title{A novel auxiliary equation neural networks method for exactly explicit solutions of nonlinear partial differential equations}
\author[label1,label2]{Shanhao Yuan}
\author[label1]{Yanqin Liu\corref{cor}} 
\ead{yqliumath@163.com}
\author[label3,label4]{Runfa Zhang\corref{cor}}
\cortext[cor]{Corresponding author.}
\ead{rf_zhang@sina.cn}
\author[label1]{Limei Yan}
\author[label1]{Shunjun Wu}
\author[label5]{Libo Feng}

\affiliation[label1]{ 
	organization={School of Mathematics and Big Data},
	addressline={Dezhou University}, 
	postcode={Dezhou 253023}, 	        
	country={China}
}

\affiliation[label2]{
	organization={School of Mathematics and Statistics},
	addressline={Qilu University of Technology (Shandong Academy of Sciences)}, 
	postcode={Jinan 250353}, 	        
	country={China}
}

\affiliation[label3]{
	organization={School of Automation and Software Engineering},
	addressline={Shanxi University}, 
	postcode={Taiyuan 030013}, 	        
	country={China}
}

\affiliation[label4]{
	organization={Hubei Key Laboratory of Applied Mathematics},
	addressline={Hubei University}, 
	postcode={Wuhan 430062}, 	        
	country={China}
}

\affiliation[label5]{
	organization={School of Mathematical Sciences},
	addressline={Queensland University of Technology}, 
	postcode={GPO Box 2434, Brisbane, Qld. 4001}, 	        
	country={Australia}
}
\begin{abstract}
In this study, we firstly propose an auxiliary equation neural networks method (AENNM), an innovative analytical method that integrates neural networks (NNs) models with the auxiliary equation method to obtain exact solutions of nonlinear partial differential equations (NLPDEs). A key novelty of this method is the introduction of a novel activation function derived from the solutions of the Riccati equation, establishing a new mathematical link between differential equations theory and deep learning. By combining the strong approximation capability of NNs with the high precision of symbolic computation, AENNM significantly enhances computational efficiency and accuracy.  To demonstrate the effectiveness of the AENNM in solving NLPDEs, three numerical examples are investigated, including the nonlinear evolution equation, the Korteweg-de Vries-Burgers equation, and the (2+1)-dimensional Boussinesq equation. Furthermore, some new trial functions are constructed by setting specific activation functions within the "2-2-2-1" and "3-2-2-1" NNs models. By embedding the auxiliary equation method into the NNs framework, we derive previously unreported solutions. The exact analytical solutions are expressed in terms of hyperbolic functions, trigonometric functions, and rational functions. Finally, three-dimensional plots, contour plots, and density plots are presented to illustrate the dynamic characteristics of the obtained solutions. This research provides a novel methodological framework for addressing NLPDEs,  with broad applicability  across scientific and engineering fields.
\end{abstract}

\end{frontmatter}



\section{Introduction}
Nonlinear partial differential equations (NLPDEs) are important mathematical tools for describing complex natural phenomena and engineering problems. Unlike their linear counterparts, NLPDEs exhibit intricate behaviors such as blow-up, solitons, shock waves, and chaos \cite{Sang2024Soliton, chen2024Global, wang2024chaos}, making them both challenging and essential for understanding real-world systems. Compared to numerical methods, analytical solutions not only reveal the intrinsic symmetries and conservation laws of a system but also serve as benchmark validations for numerical simulations. For instance, in optical fiber communications, the soliton solutions of the nonlinear Schr\"odinger equation provide the theoretical foundation for distortion-free pulse propagation \cite{Roshid2024Bifurcation}, while the exact solutions of shallow water wave equations are critical for tsunami early-warning modeling  \cite{Yousaf2025wave}.
Exploring exact solutions to NLPDEs has become a key focus in understanding various physical phenomena and continues to attract growing attention from researchers worldwide. 

In the past several decades,
many analytical methods have been established and developed. 
Within the integrability framework, Hirota's bilinear method \cite{zhang2024study, Yang2024the} transforms the nonlinear equation into a bilinear form through variable substitution, thereby enabling the direct construction of soliton solutions. Complementary approaches like the B\"acklund transformation \cite{wu2022backlund, liu2024lump, manda2024integrability} establish differential constraints between solutions, providing a generative mechanism for exotic localized waves such as lump solitons. As a classical tool for integrable systems, the inverse scattering method \cite{sasaki2024on, ali2023travelling, zhang2022soliton} converts nonlinear problems into linear integral equations through spectral theory, and has been successfully applied to models such as the KdV equation. For traveling wave solutions, the Jacobi elliptic function method \cite{khan2024analyzing, farooq2024exact}, leverages periodicity to approximate solutions, though its efficacy depends on specific nonlinear term structures. The tanh-function expansion and its various extension \cite{tian2019quasi, almatrafi2023solitary, ei-tantawy2021novel},  solves traveling wave solutions of nonlinear equations through polynomial combinations of hyperbolic tangent functions. The F-expansion method \cite{kaur2019dispersion, muhammad2024on, mohamed2023investigation} further broadens applicability to higher dimensions through auxiliary ordinary differential equations. Notably, the (G'/G)-expansion method \cite{ali2024two, burcu2012the, wang2008the} parametrizes solution families using rational functions of  solutions to linear ordinary differential equations, while the exp-function method \cite{liu2018multiple, yakup2017a, Ravi2017new} employs exponential terms to construct multi-peakon solutions, albeit requiring rigorous validation to exclude spurious solutions.  However, the diversity of nonlinear terms in NLPDEs, including dispersion, dissipation, and convective effects, makes a universal analytical method unattainable.

With the emergence of artificial intelligence, deep learning is widely used in the fields of science and technology \cite{li2024ldelay, choudhary2022recent, liu2024dendritic}. The universal approximation theorem states that a fully connected neural networks (NNs) can approximate any continuous function with arbitrary accuracy \cite{Hornik1989Multilayer}. Raissi et al. \cite{raissi2019physics} introduced physics-informed neural networks (PINNs), a class of machine learning models that embed physical laws directly into their training process to solve NLPDEs.
Lu et al. \cite{lu2021operators} proposed deep operator network (DeepONet), a neural network with small generalization error for learning nonlinear operators by combining a branch net to encode the discrete input function space and a trunk net to encode domain of the output functions. Subsequently, Wang et al. \cite{wang2021operators} developed physics-informed DeepONet, which first employs DeepONet to learn the solution operator of PDEs, then utilizes this operator as prior knowledge in PINNs to accelerate the training process while achieving higher accuracy. Nevertheless, these NNs methods provide approximate solutions rather than exact solutions, inevitably introducing numerical errors. Being data-driven models, these techniques necessitate large-scale training data, which imposes significant computational overhead. Moreover, the performance of NNs heavily depends on optimization algorithms for parameter tuning, which can be computationally intensive. Therefore, improving training efficiency and developing more robust optimization strategies remain critical research directions.

Recently, the integration of NNs with symbolic computation for quickly obtaining exact solutions to NLPDEs has attracted significant research attention. Zhang et al. \cite{Zhang2019Bilinear} proposed the bilinear neural network method (BNNM), which combines the bilinear approach with NNs models to obtain exact solutions for NLPDEs. Subsequently, the bilinear residual network method \cite{Zhang2022residual} was developed to obtain exact explicit solutions for nonlinear evolution equations while reducing model complexity.
Liu et al. \cite{Liu2023multivariate} investigated a multivariate bilinear neural network method by improving the BNNM, introducing multivariate functions as activation functions in the NNs. Wu et al. \cite{wu2024variable} introduced a variable coefficient bilinear residual network method, which effectively addresses the issue of variable coefficients in nonlinear integrable systems with variable coefficients.  Zhu et al. \cite{Zhu2025BellPolynomial} derived the bilinear form of the fractional (3+1)-dimensional Yu-Toda-Sasa-Fukuyama equation by combining the Bell polynomial approach with NNs. 

Inspired by the above works, an auxiliary equation neural networks method, which embeds the auxiliary equation method into the NNs architecture, is firstly proposed to solve NLPDEs in this paper. Our major contributions are highlighted as follows:
\begin{itemize}
	\item[$\bullet$] We apply NNs architecture to construct the potential analytical solution of NLPDEs. The output of AENNM, obtained through feedforward computation comprising weights, biases and activation functions, serves as a trial function for the NLPDEs.
	\item[$\bullet$] A novel activation function derived from the solutions of the Riccati equation is creatively introduced for NNs, establishing a new mathematical connection between differential equations and deep learning.
	\item[$\bullet$] We provide an innovative method that utilizes NNs without data samples to obtain exact symbolic analytical solutions of NLPDEs. Thus, the computational efficiency and accuracy of NLPDEs are significantly improved.
	\item[$\bullet$]  This approach employs NNs to systematize and clarify the construction of the trial function. The network structure introduced is both adaptable and configurable, allowing it to be tailored to diverse types of NLPDEs through modifications in the number of network layers, neurons, and types of activation functions.

\end{itemize}

The structure of this paper is as follows. In \Cref{sec:methodology}, we introduce the theory of the proposed method. In \Cref{examples}, the accuracy and feasibility of the analytical method for NLPDEs are verified through the nonlinear evolution equation, the Korteweg-de Vries-Burgers equation, and the (2+1)-dimensional 
Boussinesq equation. In \Cref{comparision}, the proposed method is compared with existing method in the literature. Finally, the discussions and conclusions of this paper are given in \Cref{discussions} and \Cref{conclusions}, respectively.
\section{Auxiliary equation neural networks method (AENNM)} \label{sec:methodology}
In this section, we introduce the idea of the AENNM
for finding solutions of NLPDEs. Consider the following general NLPDEs
\begin{equation}\label{(2.1)}
	\mathcal{L}\{u(x,t)\} + \mathcal{N}\{u(x,t)\}= 0,
\end{equation}
where $u(x,t)$ is the solution of the equation, $\mathcal{L}\{u(x,t)\}$ is a linear operator that involves $u(x,t)$ and its various partial derivatives with respect to $x$ and $t$,  $\mathcal{N}\{u(x,t)\}$ is a nonlinear operator that encompasses  $u(x,t)$ and its various partial derivatives.

The method uses the output of the NNs as a trial function to obtain the analytical solutions of the NLPDEs. We initially simplify the NLPDEs into a set of easily solvable nonlinear algebraic equations by applying the trial function. Subsequently, we determine the corresponding weights and biases of NNs by solving the system of nonlinear algebraic equations.
 
\begin{figure}[h]
	\begin{subfigure}[t]{0.5\linewidth}
		\centering
		\includegraphics[width=1\textwidth]{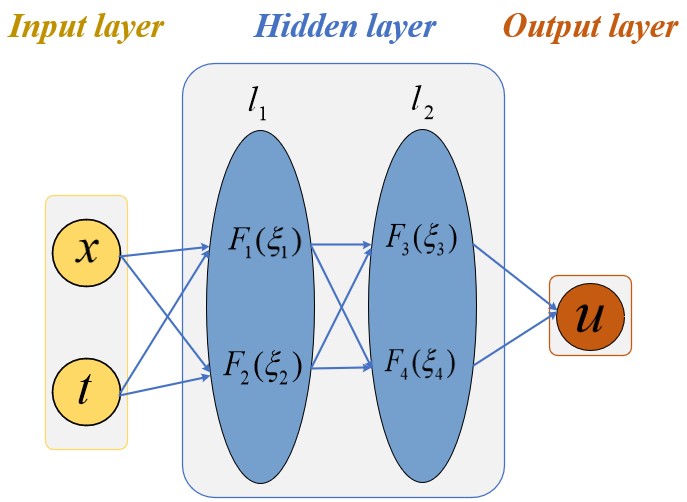}
		\subcaption{2-2-2-1 model}
	\end{subfigure}
	\hfill 
	\begin{subfigure}[t]{0.5\linewidth}
		\centering
		\includegraphics[width=1\textwidth]{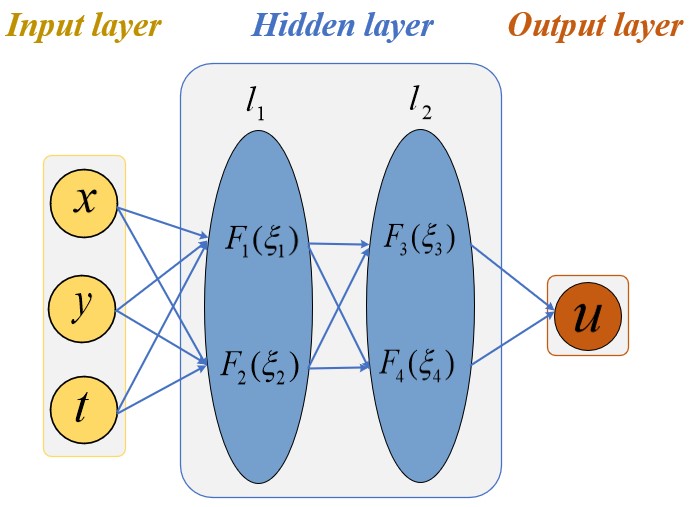}
		\subcaption{3-2-2-1 model}
	\end{subfigure}
	\caption{NNs architectures for obtaining the exact analytical solutions of NLPDEs.}
	\label{Figure_1}
\end{figure}

To obtain the solution $u(x,t)$ of the NLPDEs, we utilize a 2-2-2-1 model within the feedforward computation of the NNs, which has two hidden layers $l_{1}$ and $l_{2}$, each consisting of two neurons, as shown in \Cref{Figure_1}a. Thus, the mathematical mapping corresponding to the explicit model is as follows
\begin{equation}\label{(2.2)}
	\begin{cases}
		\xi_{1}=tw_{t1}+xw_{x1}+b_{1},\\
		\xi_{2}=tw_{t2}+xw_{x2}+b_{2},\\
		\xi_{3}=w_{23}F_{2}(\xi_{2})+w_{13}F_{1}(\xi_{1})+b_{3},\\
		\xi_{4}=w_{24}F_{2}(\xi_{2})+w_{14}F_{1}(\xi_{1})+b_{4},\\
		u(x,t)=w_{3u}F_{3}(\xi_{3})+w_{4u}F_{4}(\xi_{4})+b_{5},
	\end{cases}
\end{equation}
where $\xi_{1}$ and $\xi_{2}$ are the outputs of the first and second neurons in the first hidden layer respectively, $\xi_{3}$ and $\xi_{4}$ are the outputs of the first and second neurons in the second hidden layer respectively, $w_{ij}(i=t,x,1,2,3,4, j=1,2,3,4,u)$ are the weights between the each neurons, $F_{1}$, $F_{2}$, $F_{3}$, $F_{4}$ are the activation functions,  $b_{1}$, $b_{2}$, $b_{3}$, $b_{4}$, $b_{5}$ are the biases of each neurons. The trial function $u(x,t)$ of the \Cref{(2.2)} based on NNs architecture is composed of weights and biases.

Unlike the 2-2-2-1 model, the 3-2-2-1 model has three inputs while keeping the rest of the networks structure unchanged, as shown in \Cref{Figure_1}b. Thus, the trial function $u(x,y,t)$ is

\begin{equation}\label{(2.3)}
	\begin{cases}
		\xi_{i}=tw_{ti}+yw_{yi}+xw_{xi}+b_{i},(i=1,2),\\
		\xi_{j}=w_{2j}F_{2}(\xi_{2})+w_{1j}F_{1}(\xi_{1})+b_{j}, (j=3,4),\\
		u(x,y,t)=w_{3u}F_{3}(\xi_{3})+w_{4u}F_{4}(\xi_{4})+b_{5}.
	\end{cases}
\end{equation}

In this paper, a novel activation function
\begin{equation}\label{(2.4)}
	\varphi(\xi_{i})=\begin{cases}
		-\sqrt{-b}\tanh\sqrt{-b}\,\xi_{i},\;\:\, b<0,\\
		
		\vspace{2pt}
		-\sqrt{-b}\coth\sqrt{-b}\,\xi_{i},\;\;\, b<0, \\
		
		\vspace{3pt}
		-\frac{1}{\xi_{i}}, \quad   b=0, \\
		\sqrt{b}\tan\sqrt{b}\,\xi_{i},\quad\quad\; b>0,\\
		-\sqrt{b}\cot\sqrt{b}\,\xi_{i}, \quad\;\;b>0, 		
	\end{cases}
\end{equation}
derived from the solutions of the Riccati equation
\begin{equation}\label{(2.5)}
	\varphi'=b+\varphi^2,
\end{equation}
is creatively introduced for NNs.

In the AENNM, if the activation functions 
 $F_{1}$ and $F_{2}$ in the first hidden layer are specified as 
$\varphi(\cdot)$, the solution of the NLPDEs can be expressed as a polynomial of 
$\varphi(\cdot)$. The proposed method provides interpretable activation functions derived from the auxiliary equation, thereby eliminating the blindness in activation function selection. The AENNM merges the precision characteristic of symbolic computation with the adaptive capabilities inherent in NNs, substantially enhancing both computational efficiency and accuracy.
\begin{figure}[H]
	\centering
	\includegraphics[width=0.8\textwidth]{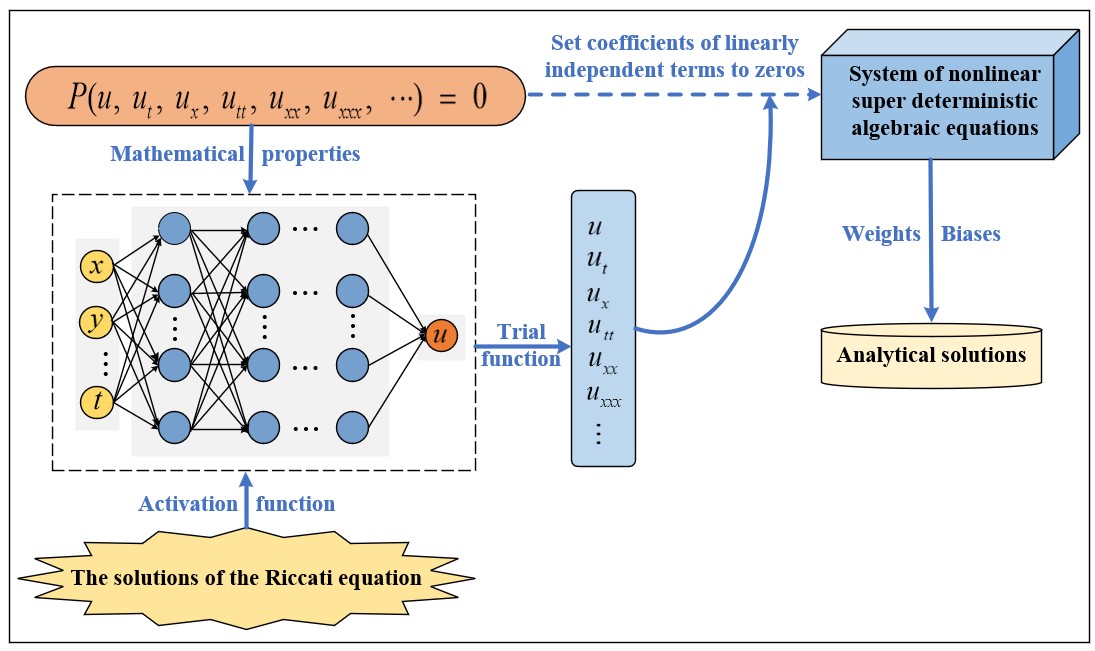}
	\caption{The algorithm flow chart of AENNM.}
	\label{Figure_2}
\end{figure}
Generally, the algorithm flow chart of AENNM is presented in \Cref{Figure_2} and the primary steps for employing the AENNM to solve the NLPDEs are as follows:

{\bf Step 1:} The Riccati equation \Cref{(2.5)} is selected as the auxiliary equation, and its solution is employed as the activation function $\varphi(\cdot)$ in the NNs.

{\bf Step 2:} The NNs model is constructed by selecting specific activation functions, the number of neurons, and the number of hidden layers, as shown in \Cref{Figure_1}.

{\bf Step 3:} A nonlinear algebraic equation governing the weights and biases is systematically derived via substitution of the trial function specified in \Cref{(2.2)} into the NLPDEs \Cref{(2.1)}.

{\bf Step 4:} By collecting coefficients of each term and equating them to zero, an underdetermined system of nonlinear algebraic equations is obtained.

{\bf Step 5:} The derived system of nonlinear algebraic equations is computationally solved employing Maple symbolic computation software.

{\bf Step 6:} By substituting the obtained weights and biases along with the specific form \Cref{(2.4)} of $\varphi(\cdot)$ into NNs model \Cref{(2.2)}, the analytical solution of \Cref{(2.1)} is derived.

\section{Examples}\label{examples}
In this section, several validation examples are considered
to verify the effectiveness of auxiliary equation neural networks method for solving the NLPDEs.

\vspace{3pt}
{\bf Example 1.} Consider  the nonlinear evolution equation \cite{fan2000extended, elwakil2002modified} of the following form
\begin{equation}\label{(3.1)}
	u_{tt}+\alpha u_{xx}+\beta u + \gamma u^3=0,
\end{equation}
where $\alpha$, $\beta$, and $\gamma$ are constants. \Cref{(3.1)} contains some particular important equations such as Duffing, Klein–Gordon, Landau–Ginsburg–Higgs, and $\phi ^{4}$ equations.

The 2-2-2-1 model is used to construct the potential analytical solution of \Cref{(3.1)}, as shown in  \Cref{Figure_1}a. By setting  $F_{1}(\xi_{1})=\varphi(\xi_{1}), F_{2}(\xi_{2})=\varphi^{2}(\xi_{2}), F_{3}(\xi_{3})=(\xi_{3}),$ and $F_{4}(\xi_{4})=(1/\xi_{4})$ in \Cref{Figure_1}a, and substituting them into \Cref{(2.2)}, we obtain the following trial function
\begin{equation}\label{(3.2)}
	\begin{aligned}  
		u(x,t)=&w_{3u} \left(w_{13} \varphi \! \left(t w_{t1}+x w_{x 1}+b_{1}\right)+w_{23} \varphi^{2} \! \left(t w_{t2}+x w_{x 2}+b_{2}\right)+b_{3}\right)\\
		&+\frac{w_{4u}}{w_{14} \varphi \! \left(t w_{t1}+x w_{x 1}+b_{1}\right)+w_{24} \varphi^{2} \! \left(t w_{t2}+x w_{x 2}+b_{2}\right)+b_{4}}+b_{5}
		.
	\end{aligned}  
\end{equation}

By substituting \Cref{(3.2)} into \Cref{(3.1)}, the solution can be derived using AENNM as follows,
\begin{equation}\label{(3.3)}
	\left\{
	\begin{array}{l}
		\alpha = 
		-\frac{32 b^{3} w_{24}^{2} w_{t2}^{2}+\gamma  w_{4u}^{2}}{32 b^{3} w_{24}^{2} w_{x2}^{2}}
		, 
		\beta = -\frac{\gamma  w_{4u}^{2}}{8 b^{2} w_{24}^{2}},
		b = b, 
		b_{3} = b_{3},
		b_{4} = -b w_{24},
		b_{5} = \frac{w_{4u}}{2 b w_{24}},\\
		w_{13} = 0,
		w_{14} = 0,
		w_{23} = w_{23},
		w_{24} = w_{24},
		w_{3u} = 0,
		w_{4u} = w_{4u},
		w_{t1} = w_{t1},\\ 
		w_{t2} = w_{t2},
		w_{x1} = w_{x1}, 
		w_{x2} = w_{x2}.		
	\end{array}
	\right\}
\end{equation}

Substituting \Cref{(3.3)} into \Cref{(3.2)} yields the following analytical solution to the original equation,
\begin{equation}\label{(3.4)}
	\begin{aligned}  
		u(x,t)=&\frac{w_{4u}}{w_{24} \varphi^{2} \! \left(t w_{t2}+x w_{x 2}+b_{2}\right)-b w_{24}}+\frac{w_{4u}}{2 b w_{24}}
		.
	\end{aligned}  
\end{equation}

When $b<0$, 
\begin{equation}\label{(3.5)}
	\begin{aligned}  
		u(x,t)=&\mp  \frac{w_{4,u}}{2 b w_{24} \cosh \! \left(2 \sqrt{-b}\, \left(t w_{t2}+x w_{x2}+b_{2}\right)\right)}.
	\end{aligned}  
\end{equation}

\begin{figure}[H]
	\centering
	\begin{subfigure}[b]{.48\linewidth}
		\centering
		\includegraphics[scale=0.5]{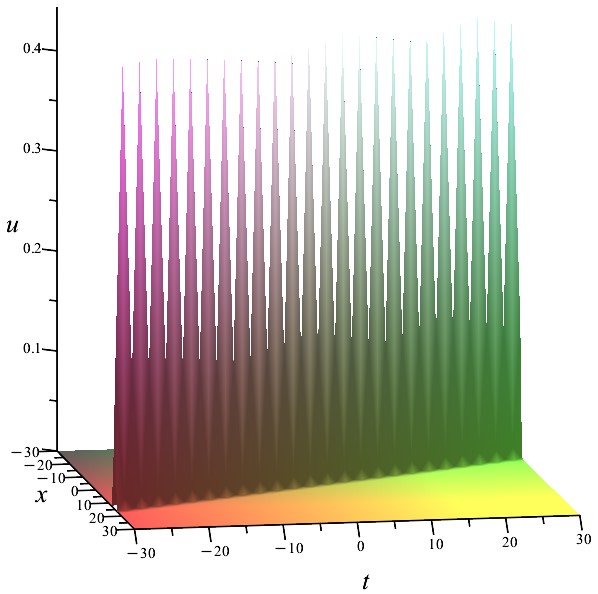}
		\subcaption{3D plot}
	\end{subfigure}
	\hfill 
	\begin{subfigure}[b]{.48\linewidth}
		\centering
		\includegraphics[scale=0.5]{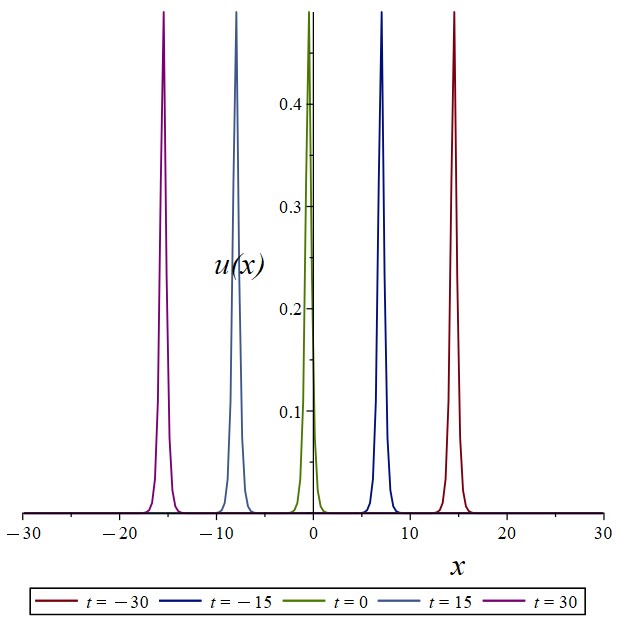}
		\subcaption{$x$-curves}
	\end{subfigure}
	\vspace{1em} 
	\begin{subfigure}[b]{.48\linewidth}
		\centering
		\includegraphics[scale=0.5]{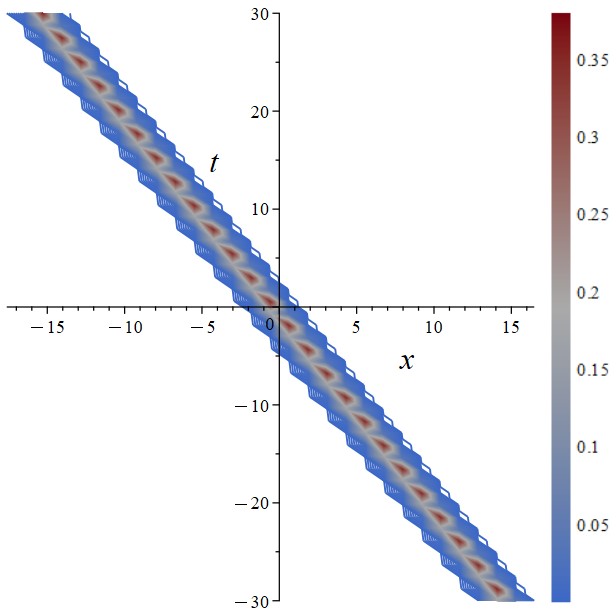}
		\subcaption{Contour plot}
	\end{subfigure}
	\hfill
	\begin{subfigure}[b]{.48\linewidth}
		\centering
		\includegraphics[scale=0.5]{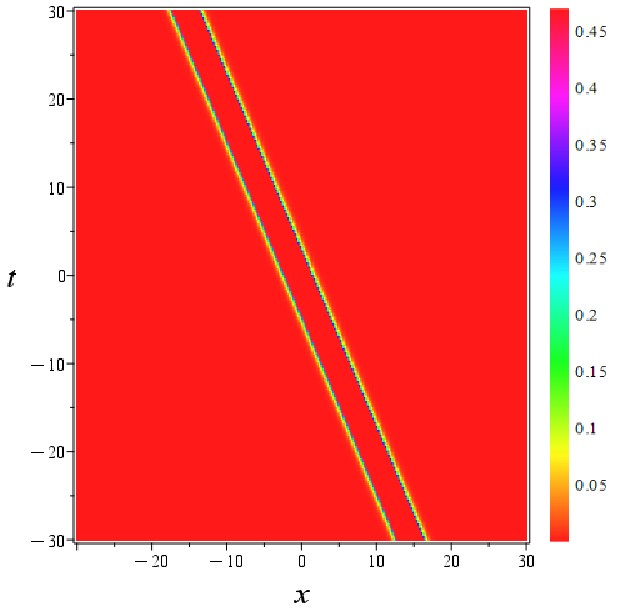}		
		\subcaption{Density plot}
	\end{subfigure}
	\caption{The three-dimensional plot, curves plot, contour plot, and density plot of the solution \Cref{(3.5)}.}
	\label{Figure_3}
\end{figure}

If one chooses coefficients $\{b=-1, w_{x2} = 2,  w_{t2} = 1, w_{24}=2, w_{4u}=1, b_{2}=1\}$ and take the "$-$" in the \Cref{(3.5)}, we obtain \Cref{Figure_3}. Then the behavior of the solution can be intuitively expressed. \Cref{Figure_3}a is a three-dimensional diagram drawn based on the spatio-temporal domain $[-30, 30] \times [-30, 30]$. To observe the local  behavior of the solution, we plot the $x$-curves in \Cref{Figure_3}b at interval $t \in [-30, 30]$. And the contour and density plots of the solution are shown in \Cref{Figure_3}b and \Cref{Figure_3}c, respectively.

When $b>0$,
\begin{equation}\label{(3.6)}
	\begin{aligned}  
		u(x,t)=&\mp \frac{w_{4u}}{2  b w_{24} \cos \! \left(2 \sqrt{b}\, \left(t w_{t2}+x w_{x2}+b_{2}\right)\right)}.
	\end{aligned}  
\end{equation}

\begin{figure}[H]
	\centering
	\begin{subfigure}[b]{.48\linewidth}
		\centering
		\includegraphics[scale=0.5]{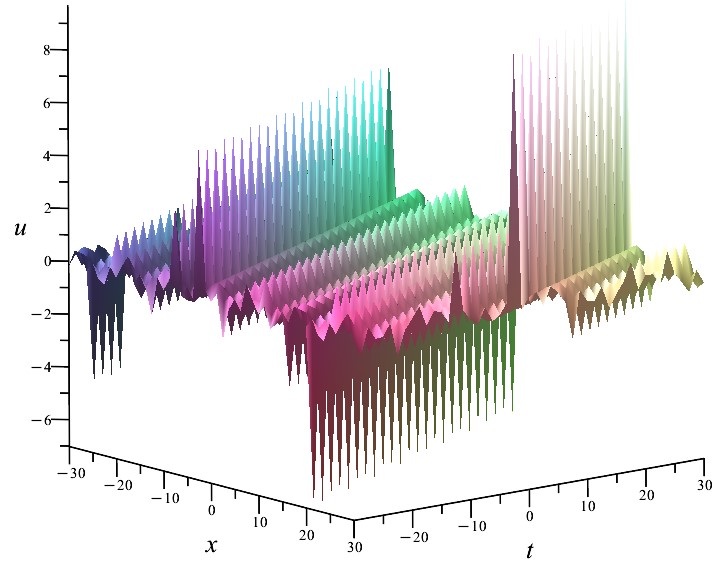}
		\subcaption{3D plot}
	\end{subfigure}
	\hfill 
	\begin{subfigure}[b]{.48\linewidth}
		\centering
		\includegraphics[scale=0.5]{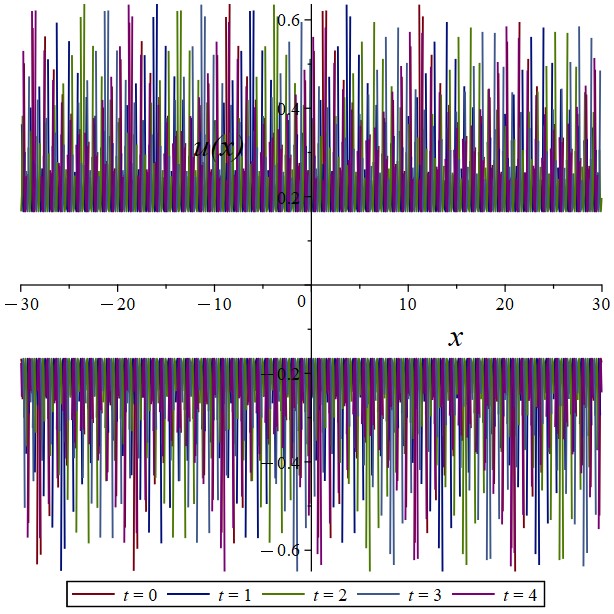}
		\subcaption{$x$-curves}
	\end{subfigure}
	\vspace{1em} 
	\begin{subfigure}[b]{.48\linewidth}
		\centering
		\includegraphics[scale=0.5]{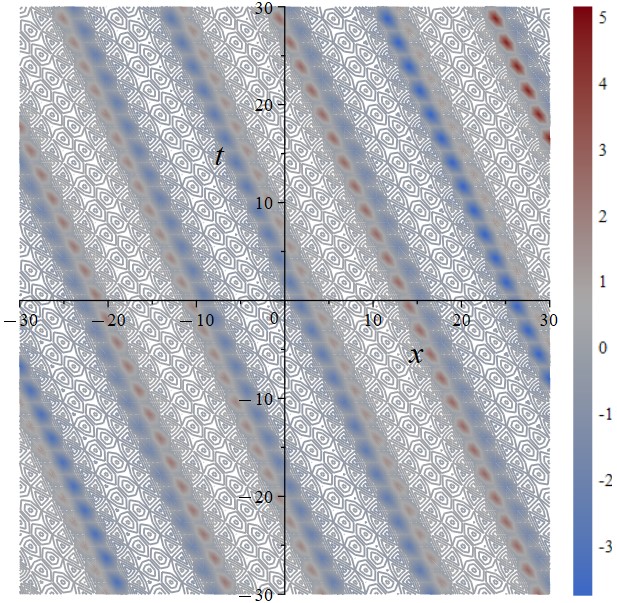}
		\subcaption{Contour plot}
	\end{subfigure}
	\hfill
	\begin{subfigure}[b]{.48\linewidth}
		\centering
		\includegraphics[scale=0.5]{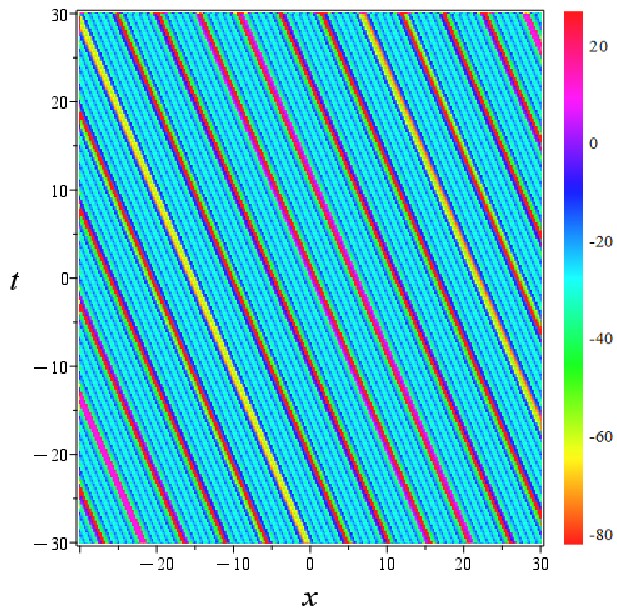}		
		\subcaption{Density plot}
	\end{subfigure}
	\caption{The three-dimensional plot, curves plot, contour plot, and density plot of the solution \Cref{(3.6)}.}
	\label{Figure_4}
\end{figure}

Similarly, to better observe the properties of the solution, we take the coefficients $\{b=3, w_{x2} = 2,  w_{t2} = 1, w_{24}=1, w_{4u}=1, b_{2}=1\}$ and take the "$-$" in the \Cref{(3.6)}. The three-dimensional plot, contour plot, and density plot of the solution are displayed in \Cref{Figure_4} based on the spatio-temporal domain $[-30, 30] \times [-30, 30]$.

{\bf Example 2.} Consider the Korteweg-de Vries-Burgers equation \cite{inc2005extended} in the form
\begin{equation}\label{(3.7)}
	u_{t}+puu_{x}+qu^2u_{x}+ru_{xx}-su_{xxx}=0,
\end{equation}
where $p$, $q$, $r$, and $s$ are constants. \Cref{(3.7)} can be regarded as a generalization of the Korteweg-de Vries (KdV), modified Korteweg-de Vries (mKdV), and Burgers equations, encompassing both nonlinear dispersion and dissipation effects.

The 2-2-2-1 model is used to construct the potential analytical solution of \Cref{(3.7)}, as shown in  \Cref{Figure_1}a. By setting  $F_{1}(\xi_{1})=\varphi(\xi_{1}), F_{2}(\xi_{2})=\varphi(\xi_{2}), F_{3}(\xi_{3})=(\xi_{3}),$  $F_{4}(\xi_{4})=(1/\xi_{4})$ in \Cref{Figure_1}a, and substituting them into \Cref{(2.2)}, we obtain the following trial function
\begin{equation}\label{(3.8)}
	\begin{aligned}  
		u(x,t)=&w_{3u} \left(w_{13} \varphi \! \left(t w_{t1}+x w_{x 1}+b_{1}\right)+w_{23} \varphi \! \left(t w_{t2}+x w_{x 2}+b_{2}\right)+b_{3}\right)\\
		&+\frac{w_{4u}}{w_{14} \varphi \! \left(t w_{t1}+x w_{x 1}+b_{1}\right)+w_{24} \varphi \! \left(t w_{t2}+x w_{x 2}+b_{2}\right)+b_{4}}+b_{5}.
	\end{aligned}  
\end{equation}

By substituting \Cref{(3.8)} into \Cref{(3.7)}, the following coefficient solutions are obtained via AENNM,

{\bf Solution 1:}
\begin{equation}\label{(3.9)}
	\left\{
	\begin{array}{l}
		p = p, 
		q = q,
		r = r,
		s = \frac{w_{23}^{2}  w_{3u}^{2} q}{6 w_{x2}^{2}},
		b = b, 	
		b_{3} = 
		\frac{-w_{3u} \left(2 q b_{5}+p \right) w_{23}-2 r w_{x 2}}{2 q w_{23} w_{3u}^{2}}
		,
		b_{4} = 0, \\
		b_{5} = b_{5},
		w_{13} = 0,
		w_{14} = w_{14},
		w_{23} = w_{23},
		w_{24} = 0,
		w_{3u} = w_{3u},
		w_{4u} = 0,\\
		w_{t1} = w_{t1}, 
		w_{x1} = w_{x1}, 
		w_{x2} = w_{x2},
		w_{t2} = \frac{w_{x2} \left(4 b \,q^{2} w_{23}^{4} w_{3u}^{4}+3 p^{2} w_{23}^{2} w_{3u}^{2}-12 r^{2} w_{x 2}^{2}\right)}{12 w_{23}^{2} q w_{3u}^{2}}.
	\end{array}
	\right\}
\end{equation}

By substituting \Cref{(3.9)} into \Cref{(3.8)}, the analytical solution of the original equation is obtained as follows,
\begin{equation}\label{(3.10)}
	\begin{aligned}  
		u(x,t)=&w_{3u} w_{23} \varphi \! \left(\frac{t w_{x2} \left(4 b \,q^{2} w_{23}^{4} w_{3u}^{4}+3 p^{2} w_{23}^{2} w_{3u}^{2}-12 r^{2} w_{x2}^{2}\right)}{12 w_{23}^{2} q w_{3u}^{2}}+x w_{x 2}+b_{2}\right)\\
		&-\frac{p}{2 q}-\frac{r w_{x2}}{q w_{23} w_{3u}}
		.
	\end{aligned}  
\end{equation}

When $b<0$, 
\begin{subequations}\label{(3.11)}
	\begin{align}  
		u(x,t) = &-w_{3u} w_{23} \sqrt{-b}\, \tanh \! \left(\sqrt{-b}\, \left(t w_{t2}+x w_{x 2}+b_{2}\right)\right)-\frac{p}{2 q}-\frac{r w_{x 2}}{q w_{23} w_{3u}}, \label{(3.11a)}\\
		u(x,t) = & -w_{3u} w_{23} \sqrt{-b}\, \coth \! \left(\sqrt{-b}\, \left(t w_{t2}+x w_{x 2}+b_{2}\right)\right)-\frac{p}{2 q}-\frac{r w_{x 2}}{q w_{23} w_{3u}}, \label{(3.11b)}
	\end{align}  
\end{subequations}
where $w_{t2}$ is defined in \Cref{(3.9)}. 

\vspace{2pt}
When $b>0$, 
\begin{subequations}\label{(3.12)}
	\begin{align}  
		u(x,t) = \,&w_{3u} w_{23} \sqrt{b}\, \tan \! \left(\sqrt{b}\, \xi_{2}\right)-\frac{p}{2 q}-\frac{r w_{x 2}}{q w_{23} w_{3u}},\\
		u(x,t) =\, &-w_{3u} w_{23} \sqrt{b}\, \cot \! \left(\sqrt{b}\, \left(t w_{t2}+x w_{x 2}+b_{2}\right)\right)-\frac{p}{2 q}-\frac{r w_{x 2}}{q w_{23} w_{3u}},
	\end{align}
\end{subequations}
where $w_{t2}$ is defined in \Cref{(3.9)}. 

\vspace{2pt}
When $b=0$, 
\begin{equation}\label{(3.13)}
	\begin{aligned}  
		u(x,t)=-\frac{w_{3u} w_{23}}{t w_{t2}+x w_{x 2}+b_{2}}-\frac{p}{2 q}-\frac{r w_{x2}}{q w_{23} w_{3u}},
	\end{aligned}  
\end{equation}
where $w_{t2}$ is defined in \Cref{(3.9)}. 

\vspace{2pt}
{\bf Solution 2:}
\begin{equation}\label{(3.14)}
	\left\{
	\begin{array}{l}
		p = 
		-\frac{2 w_{x1} \left(b w_{14}^{2}+b_{4}^{2}\right) \left(6 b s b_{5} w_{14}^{2} w_{x1}-r w_{14} w_{4u}+6 s b_{4} w_{x1} \left(b_{4} b_{5}+w_{4u}\right)\right)}{w_{4u}^{2} w_{14}^{2}}
		, 
		q = \frac{6 s w_{x1}^{2} \left(b w_{14}^{2}+b_{4}^{2}\right)^{2}}{w_{4u}^{2} w_{14}^{2}}
		,\\
		r = r,
		s = s,
		b = b, 	
		b_{3}  = b_{3},
		b_{4}  = b_{4}, 
		b_{5}  = b_{5},
		w_{13} = w_{24} = w_{3u} = 0,\\
		w_{14} = w_{14},
		w_{23} = w_{23},
		w_{4u} = w_{4u}, 
		w_{x1} = w_{x1}, 
		w_{x2} = w_{x2},
		w_{t2} = w_{t2},\\
		w_{t1} = 
		\frac{\left(6 b^{2} s b_{5}^{2} w_{14}^{4} w_{x1}-2 b r b_{5} w_{14}^{3} w_{4u}+\left(12 b_{4}^{2} b_{5}^{2}+12 b_{4} b_{5} w_{4u}+2 w_{4u}^{2}\right) s b w_{x1} w_{14}^{2}-\Omega_{1}\right) w_{x 1}^{2}}{w_{4u}^{2} w_{14}^{2}},\\
		\Omega_{1} =2 r b_{4} w_{4u} \left(b_{4} b_{5}+w_{4u}\right) w_{14}+6 s b_{4}^{2} w_{x1} \left(b_{4} b_{5}+w_{4u}\right)^{2}.
	\end{array}
	\right\}
\end{equation}

By substituting \Cref{(3.14)} into \Cref{(3.8)}, the solution of the original equation is obtained as follows,
\begin{equation}\label{(3.15)} 
		u(x,t)=\frac{w_{4u}}{w_{14}\, \varphi \! \left(t w_{t1}+x w_{x1}+b_{1}\right)+b_{4}}+b_{5}, 
\end{equation}
where $w_{t1}$ is defined in \Cref{(3.14)}. 

When $b<0$, 
\begin{subequations}\label{(3.16)}
	\begin{align} 
		u(x,t)=\, &\frac{w_{4u}}{-w_{14} \sqrt{-b}\, \tanh \! \left(\sqrt{-b}\, \left(t w_{t1}+x w_{x 1}+b_{1}\right)\right)+b_{4}}+b_{5},\\ 
		u(x,t) =\, &\frac{w_{4u}}{-w_{14} \sqrt{-b}\, \coth \! \left(\sqrt{-b}\, \left(t w_{t1}+x w_{x 1}+b_{1}\right)\right)+b_{4}}+b_{5}, 
	\end{align}
\end{subequations}
where $w_{t1}$ is defined in \Cref{(3.14)}.

When $b>0$, 
\begin{subequations}\label{(3.17)}
	\begin{align}  
		u(x,t) = \, &\frac{w_{4u}}{w_{14} \sqrt{b}\, \tan \! \left(\sqrt{b}\, \xi_{1}\right)+b_{4}}+b_{5},\\ 
		u(x,t) = \, &\frac{w_{4u}}{-w_{14} \sqrt{b}\, \cot \! \left(\sqrt{b}\, \left(t w_{t1}+x w_{x 1}+b_{1}\right)\right)+b_{4}}+b_{5}, 
	\end{align}
\end{subequations}
where $w_{t1}$ is defined in \Cref{(3.14)}.

When $b=0$, 
\begin{equation}\label{(3.18)}
	u(x,t)=\frac{w_{4u}}{-\frac{w_{14}}{t w_{t 1}+x w_{x 1}+b_{1}}+b_{4}}+b_{5},
\end{equation}
where $w_{t1}$ is defined in \Cref{(3.14)}.

\vspace{2pt}
{\bf Solution 3:}
\begin{equation}\label{(3.19)}
	\left\{
	\begin{array}{l}
		b = 0, 	
		p = p, 
		q = q,
		r  = 
		-\frac{w_{23} w_{3u} \left(2 q b_{3} w_{3u}+2 q b_{5}+p \right)}{2 w_{x2}},
		s = \frac{w_{23}^{2} q w_{3u}^{2}}{6 w_{x2}^{2}},
		b_{3} = b_{3},
		b_{4} = b_{4},\\ 
		b_{5} = b_{5},
		w_{13} = w_{14} = w_{4u}= 0,
		w_{23} = w_{23},
		w_{24} = w_{24},
		w_{3u} = w_{3u},
		w_{t1} = w_{t1},\\ 
		w_{x1} = w_{x1},
		w_{x2} = w_{x2},
		w_{t2} = -\left(b_{3} w_{3u}+b_{5}\right) w_{x2} \left(\left(b_{3} w_{3u}+b_{5}\right) q +p \right).
	\end{array}
	\right\}
\end{equation}

By substituting \cref{(3.19)} into \cref{(3.8)} the solution of the \cref{(3.7)} can ultimately be obtained as:
\begin{equation}\label{(3.20)}
		u(x,t)= \frac{w_{23} w_{3u}}{w_{x2} \left(\left(b_{3} w_{3u}+b_{5}\right) \left(q b_{3} w_{3u}+q b_{5}+p \right) t -x \right)}+b_{3} w_{3u}+b_{5}, 
\end{equation}
where $w_{t2}$ is defined in \Cref{(3.19)}.

\vspace{2pt}
{\bf Solution 4:}
\begin{equation}\label{(3.21)}
	\left\{
	\begin{array}{l}
		p = 
		\frac{-2 q  \left(b_{3} w_{3u}+b_{5}\right) w_{4u}+2 b r  w_{24} w_{x2}}{w_{4u}}, 
		\beta = \beta,
		q = q, 
		s = \frac{q  w_{4u}^{2}}{6 b^{2} w_{24}^{2} w_{x 2}^{2}},
		b_{3} = b_{3},
		b_{4} = 0, 
		b_{5} = b_{5},\\
		w_{13} = 0,
		w_{14} = 0,
		w_{23} = -\frac{w_{4u}}{b w_{24} w_{3u}},
		w_{24} = w_{24},
		w_{3u} = w_{3u},
		w_{4u} = w_{4u},
 		w_{t1} = w_{t1}, \\
		w_{x1} = w_{x1}, 
		w_{x2} = w_{x2},
		w_{t2} = 
		-\frac{\left(6b^{2} r  w_{x2} \left(b_{3} w_{3u}+b_{5}\right) w_{24}^{3}-3b q  w_{4u} \left(b_{3} w_{3u}+b_{5}\right)^{2} w_{24}^{2}-4 q  w_{4u}^{3}\right) w_{x2}}{3b w_{24}^{2} w_{4u}}.
	\end{array}
	\right\}
\end{equation}

The solution $u(x, t)$ corresponding to this set of constraints is as follows,
\begin{equation}\label{(3.22)}
	u(x,t)=-\frac{w_{4u} \varphi \! \left(t w_{t2}+x w_{x2}+b_{2}\right) }{b w_{24}}+\frac{w_{4u}}{w_{24} \varphi \! \left(t w_{t2}+x w_{x2}+b_{2}\right)}+b_{3} w_{3u}+b_{5},
\end{equation}
where $w_{t2}$ is defined in \Cref{(3.21)}.

When $b<0$, 
\begin{equation}\label{(3.23)}
	\begin{aligned}  
		u(x,t)=&\frac{-w_{4u} \left(\tanh \! \left(\sqrt{-b}\, \left(t w_{t2}+x w_{x2}+b_{2}\right)\right)+\coth \! \left(\sqrt{-b}\, \left(t w_{t2}+x w_{x2}+b_{2}\right)\right)\right)}{\sqrt{-b}\, w_{24}}\\
		&+b_{3} w_{3u}+b_{5},
	\end{aligned}  
\end{equation}
where $w_{t2}$ is defined in \Cref{(3.21)}.
\begin{figure}[h]
	\centering
	\begin{subfigure}[b]{.48\linewidth}
		\centering
		\includegraphics[scale=0.5]{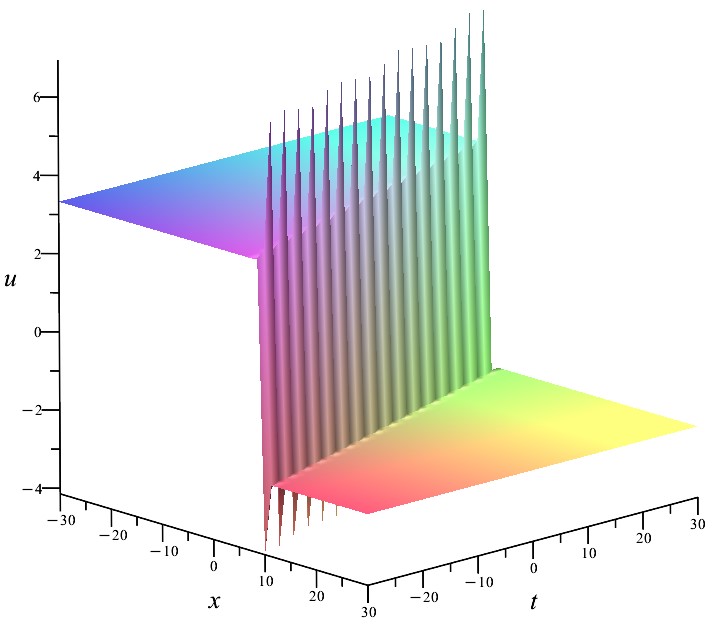}
		\subcaption{3D plot}
	\end{subfigure}
	\hfill 
	\begin{subfigure}[b]{.48\linewidth}
		\centering
		\includegraphics[scale=0.5]{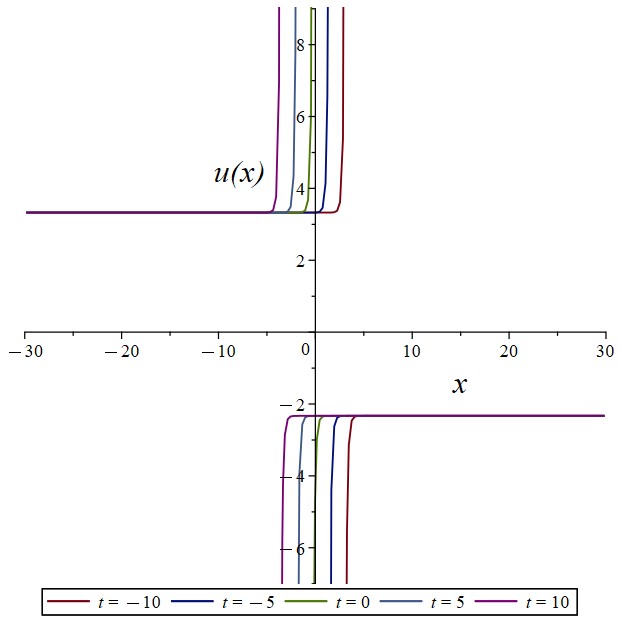}
		\subcaption{$x$-curves}
	\end{subfigure}
	\vspace{1em} 
	\begin{subfigure}[b]{.48\linewidth}
		\centering
		\includegraphics[scale=0.5]{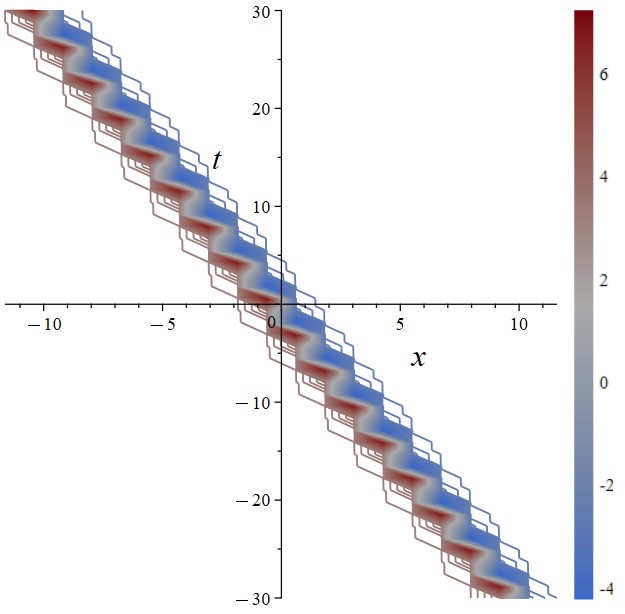}
		\subcaption{Contour plot}
	\end{subfigure}
	\hfill
	\begin{subfigure}[b]{.48\linewidth}
		\centering
		\includegraphics[scale=0.5]{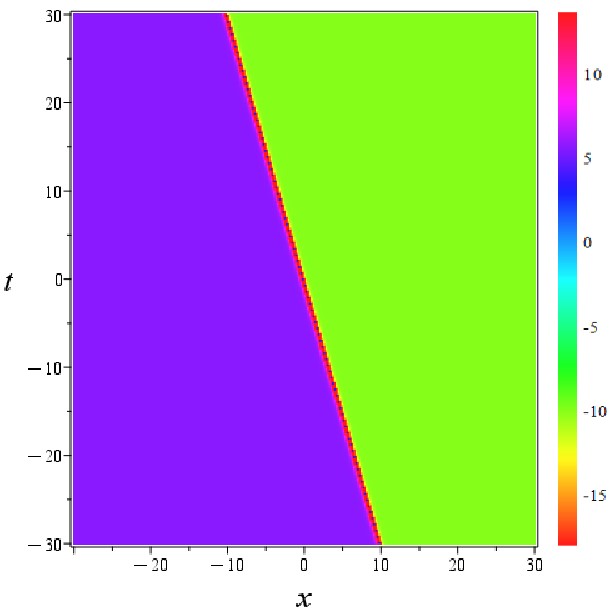}		
		\subcaption{Density plot}
	\end{subfigure}
	\caption{The three-dimensional plot, curves plot, contour plot, and density plot of the solution \Cref{(3.23)}.}
	\label{Figure_5}
\end{figure}

If one chooses coefficients $\{b=-0.5, q=1, r=1, w_{x2} = 2, w_{24}=1, w_{3u}=1, w_{4u}=1, b_{2}=0.5, b_{3}=0.5, b_{5}=0.5\}$ in the \Cref{(3.23)}, we obtain \Cref{Figure_5}. Then the behavior of the solution can be intuitively expressed. \Cref{Figure_5}a is a three-dimensional diagram drawn based on the spatio-temporal domain $[-30, 30] \times [-30, 30]$. To observe the local  behavior of the solution, we plot the $x$-curves in \Cref{Figure_5}b at interval $t \in [-30, 30]$. And the contour and density plots of the solution are shown in \Cref{Figure_5}c and \Cref{Figure_5}d, respectively.

When $b>0$, 
\begin{equation}\label{(3.24)}
	\begin{aligned}  
		u(x,t)=&\frac{w_{4u} \left(\cot \! \left(\sqrt{b}\, \left(t w_{t 2}+x w_{x2}+b_{2}\right)\right)-\tan \! \left(\sqrt{b}\, \left(t w_{t2}+x w_{x2}+b_{2}\right)\right)\right)}{\sqrt{b}\, w_{24}}\\
		&+b_{3} w_{3u}+b_{5},
	\end{aligned}  
\end{equation}
where $w_{t2}$ is defined in \Cref{(3.21)}.

\begin{figure}[H]
	\centering
	\begin{subfigure}[b]{.48\linewidth}
		\centering
		\includegraphics[scale=0.45]{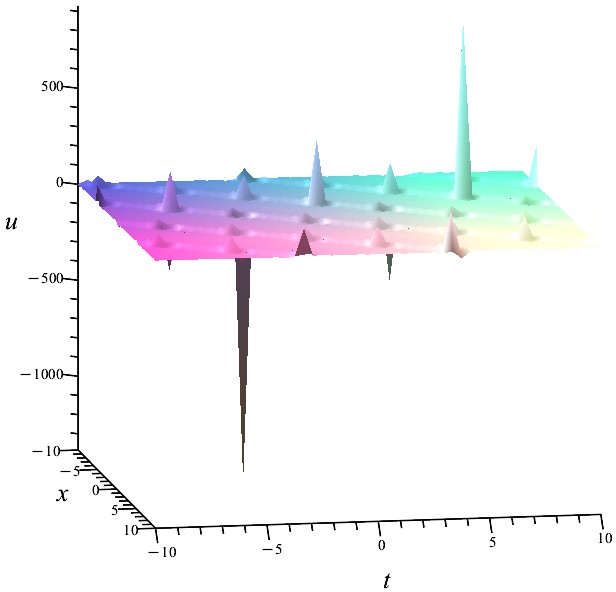}
		\subcaption{3D plot}
	\end{subfigure}
	\hfill 
	\begin{subfigure}[b]{.48\linewidth}
		\centering
		\includegraphics[scale=0.45]{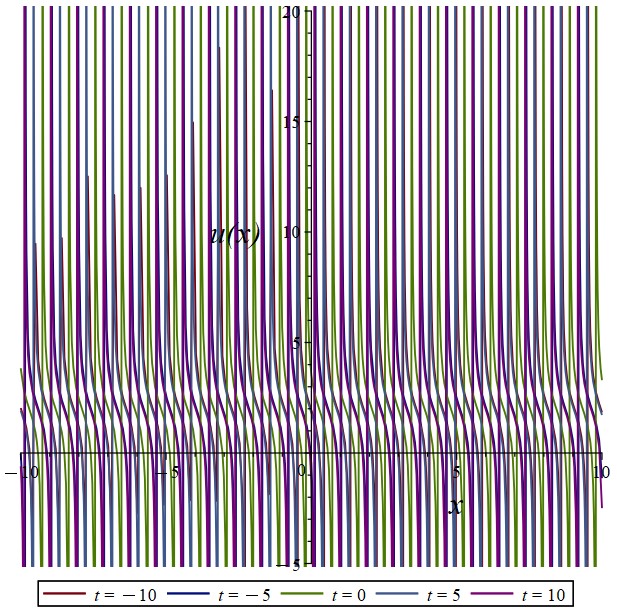}
		\subcaption{$x$-curves}
	\end{subfigure}
	\vspace{1em} 
	\begin{subfigure}[b]{.48\linewidth}
		\centering
		\includegraphics[scale=0.45]{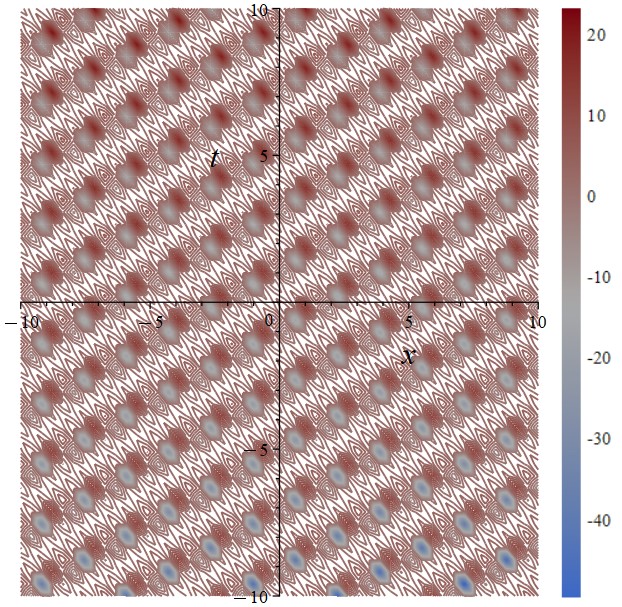}
		\subcaption{Contour plot}
	\end{subfigure}
	\hfill
	\begin{subfigure}[b]{.48\linewidth}
		\centering
		\includegraphics[scale=0.45]{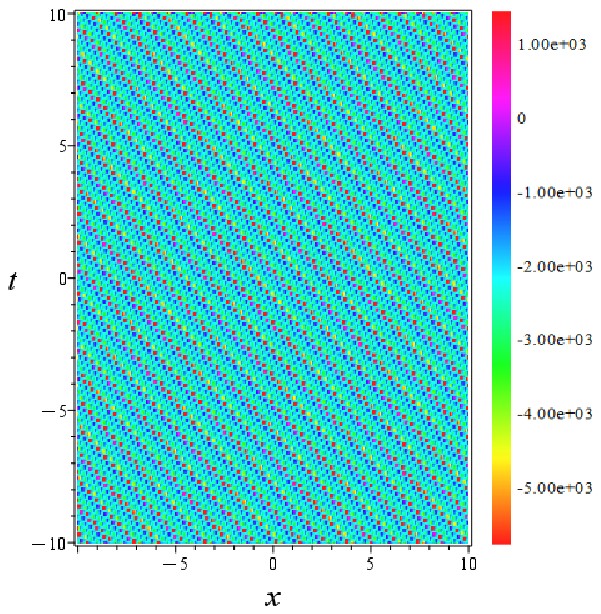}		
		\subcaption{Density plot}
	\end{subfigure}
	\caption{The three-dimensional plot, curves plot, contour plot, and density plot of the solution \Cref{(3.24)}.}
	\label{Figure_6}
\end{figure}

Similarly, to better observe the properties of the solution, we take the coefficients $\{b=3, q=1, r=1, w_{x2} = 1, w_{24}=1, w_{3u}=1, w_{4u}=1, b_{2}=2, b_{3}=1, b_{5}=1\}$  in the \Cref{(3.24)}. The three-dimensional plot, contour plot, and density plot of the solution are displayed in \Cref{Figure_6} based on the spatio-temporal domain $[-10, 10] \times [-10, 10]$.

{\bf Example 3.} Consider the (2+1)-dimensional generalization of
Boussinesq equation \cite{inc2005extended}
\begin{equation}\label{(3.25)}
	u_{tt}-u_{xx}-u_{yy}-(u^2)_{xx}-u_{xxxx}=0.
\end{equation}

The 3-2-2-1 model is used to construct the potential analytical solution of \Cref{(3.1)}, as shown in  \Cref{Figure_1}b. By setting  $F_{1}(\xi_{1})=\varphi(\xi_{1}), F_{2}(\xi_{2})=\varphi(\xi_{2}), F_{3}(\xi_{3})=(\xi_{3})^2,$ and $F_{4}(\xi_{4})=(1/\xi_{4})^2$ in \Cref{Figure_1}b, and substituting them into \Cref{(2.2)}, we obtain the following trial function

\begin{equation}\label{(3.26)}
	\begin{aligned}
	u(x,y,t) =& w_{3u} \left(w_{13} \varphi \! \left(t w_{t1}+x w_{x1}+y w_{y1}+b_{1}\right)+w_{23} \varphi \! \left(t w_{t2}+x w_{x2}+y w_{y2}+b_{2}\right)+b_{3}\right)^{2}\\
	&+\frac{w_{4u}}{\left(w_{14} \varphi \! \left(t w_{t1}+x w_{x1}+y w_{y 1}+b_{1}\right)+w_{24} \varphi \! \left(t w_{t 2}+x w_{x2}+y w_{y2}+b_{2}\right)+b_{4}\right)^{2}}\\
	&+b_{5}. 
	\end{aligned}
\end{equation}

By substituting \Cref{(3.26)} into \Cref{(3.25)}, the following coefficient solutions are obtained via AENNM,

{\bf Solution 1:}
\begin{equation}\label{(3.27)}
	\left\{
	\begin{array}{l}
		b = b, 	
		b_{3} =  b_{3},
		b_{4} = 0, 
		b_{5} = 
		\frac{-8 b w_{x2}^{4}+w_{t2}^{2}-w_{x2}^{2}-w_{y 2}^{2}}{2 w_{x2}^{2}},
		w_{13} = w_{14} = w_{4u} = 0,\\
		w_{23} = w_{23},
		w_{24} = w_{24},
		w_{3u} = -\frac{6 w_{x2}^{2}}{w_{23}^{2}},
		w_{t1} = w_{t1},
		w_{t2} = w_{t2},
		w_{x1} = w_{x1},\\
		w_{x2} = w_{x2},
		w_{y1} = w_{y1}, 
		w_{y2} = w_{y2}.
	\end{array}
	\right\}
\end{equation}

By substituting \Cref{(3.27)} into \Cref{(3.26)}, the analytical solution of the original equation is obtained as follows,
\begin{equation}\label{(3.28)} 
	u(x,t)=-6 w_{x2}^{2} \, \varphi^{2} \! \left(t w_{t2}+x w_{x2}+y w_{y 2}+b_{2}\right) -\frac{8 b w_{x2}^{4}-w_{t 2}^{2}+w_{x2}^{2}+w_{y2}^{2}}{2 w_{x2}^{2}}. 
\end{equation}

When $b<0$, 
\begin{subequations}\label{(3.29)}
	\begin{align} 
		u(x,t) = \,&6 w_{x2}^{2} b \tanh^{2}\left(\sqrt{-b}\, \xi_{2} \right)-\frac{8 b w_{x2}^{4}-w_{t 2}^{2}+w_{x2}^{2}+w_{y2}^{2}}{2 w_{x2}^{2}},\\  
		u(x,t) = \,&6 w_{x2}^{2} b \coth^{2}\left(\sqrt{-b}\, \xi_{2}\right)-\frac{8 b w_{x2}^{4}-w_{t 2}^{2}+w_{x2}^{2}+w_{y2}^{2}}{2 w_{x2}^{2}},  
	\end{align}
\end{subequations}
where $\xi_{2}=t w_{t2}+y w_{y2}+x w_{x2}+b_{2}$.

\vspace{3pt}
When $b>0$, 
\begin{subequations}\label{(3.30)}
	\begin{align} 
		u(x,t) = &-6 w_{x2}^{2} b \tan^{2}\left(\sqrt{b}\, \xi_{2}\right)-\frac{8 b w_{x 2}^{4}-w_{t2}^{2}+w_{x2}^{2}+w_{y2}^{2}}{2 w_{x2}^{2}}, \\
		u(x,t) = &-6 w_{x2}^{2} b \cot^{2}\left(\sqrt{b}\, \xi_{2}\right)-\frac{8 b w_{x2}^{4}-w_{t 2}^{2}+w_{x2}^{2}+w_{y2}^{2}}{2 w_{x2}^{2}},
	\end{align}
\end{subequations}
where $\xi_{2}=t w_{t2}+y w_{y2}+x w_{x2}+b_{2}$.

When $b=0$, 
\begin{equation}\label{(3.31)} 
	u(x,t)=-\frac{6 w_{x2}^{2}}{\left(t w_{t2}+x w_{x2}+y w_{y 2}+b_{2}\right)^{2}}+\frac{w_{t2}^{2}-w_{x 2}^{2}-w_{y2}^{2}}{2 w_{x2}^{2}}. 
\end{equation}

{\bf Solution 2:}
\begin{equation}\label{(3.32)}
	\left\{
	\begin{array}{l}
		b = b, 	
		b_{3} = 0,
		b_{4} = 0, 
		b_{5} = 
		\frac{-8 b w_{x2}^{4}+w_{t2}^{2}-w_{x2}^{2}-w_{y 2}^{2}}{2 w_{x2}^{2}},
		w_{13} = w_{14} = 0,
		w_{23} = w_{23},\\
		w_{24} = w_{24},
		w_{3u} = -\frac{6 w_{x2}^{2}}{w_{23}^{2}},
		w_{4,u} = -6 w_{x2}^{2} b^{2} w_{24}^{2},
		w_{t1} = w_{t1},
		w_{t2} = w_{t2},\\
		w_{x1} = w_{x1},
		w_{x2} = w_{x2},
		w_{y1} = w_{y1}, 
		w_{y2} = w_{y2}.
	\end{array}
	\right\}
\end{equation}

By substituting \cref{(3.32)} into \cref{(3.26)}, the solution of the \cref{(3.25)} can ultimately be obtained as
\begin{equation}\label{(3.33)} 
	\begin{aligned}
	u(x,t)=&-6 w_{x2}^{2}\, \varphi^{2} \! \left(t w_{t2}+x w_{x2}+y w_{y2}+b_{2}\right)-\frac{6 w_{x2}^{2} b^{2}}{\varphi^{2} \! \left(t w_{t2}+x w_{x2}+y w_{y2}+b_{2}\right)}\\
	&-\frac{8 b w_{x2}^{4}-w_{t2}^{2}+w_{x2}^{2}+w_{y2}^{2}}{2 w_{x 2}^{2}}.
	\end{aligned} 
\end{equation}
\begin{figure}[H]
	\centering
	\begin{subfigure}[b]{.32\linewidth} 
		\centering
		\includegraphics[scale=0.32]{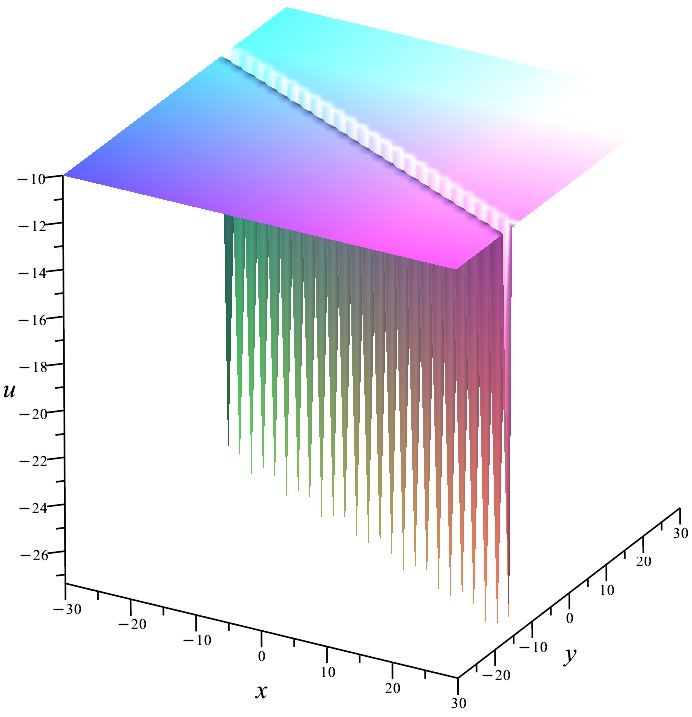}
		\subcaption{3D plot}
	\end{subfigure}
	\hfill 
	\begin{subfigure}[b]{.32\linewidth} 
		\centering
		\includegraphics[scale=0.33]{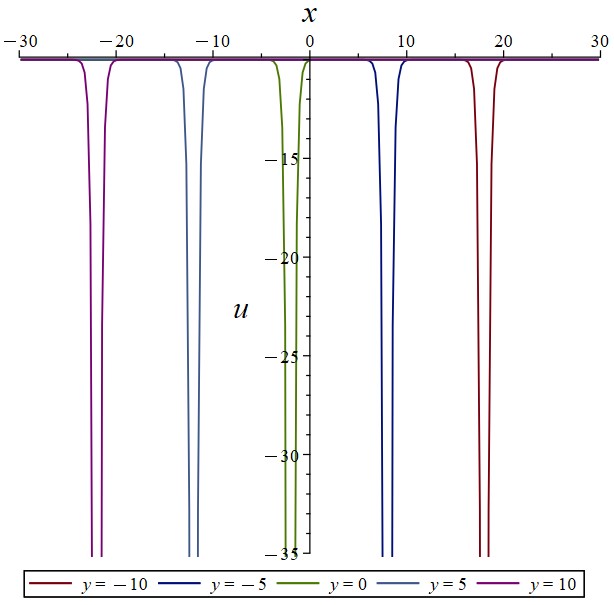}
		\subcaption{$x$-curves}
	\end{subfigure}
	\hfill 
	\begin{subfigure}[b]{.32\linewidth} 
		\centering
		\includegraphics[scale=0.33]{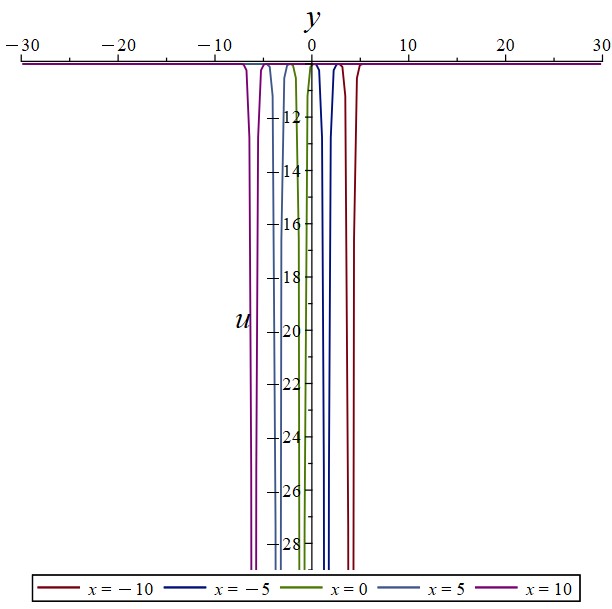}
		\subcaption{$y$-curves}
	\end{subfigure}
	
	\vspace{1em} 
	\begin{subfigure}[t]{.4\linewidth}
		\centering
		\includegraphics[scale=0.33]{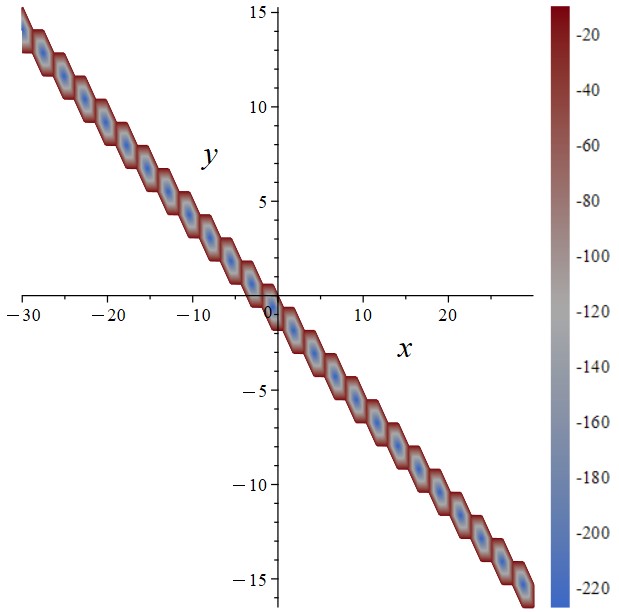}
		\subcaption{Contour plot}
	\end{subfigure}
	\hspace{-3em}
	\begin{subfigure}[t]{.4\linewidth}
		\centering
		\includegraphics[scale=0.33]{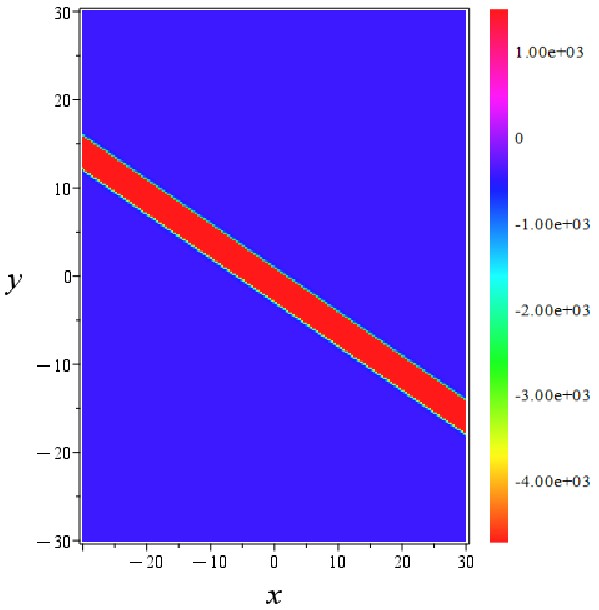}
		\subcaption{Density plot}
	\end{subfigure} 
	\centering
	\caption{The three-dimensional plot, curves plot, contour plot, and density plot of the solution \Cref{(3.34a)}.}
	\label{Figure_7}
\end{figure}
When $b<0$, 
\begin{subequations}\label{(3.34)}
	\begin{align}
		u(x,t) = \,&6 w_{x 2}^{2} b \tanh^{2}\left(\sqrt{-b}\, \xi_{2}\right) +\frac{6 w_{x2}^{2} b}{\tanh^{2} \! \left(\sqrt{-b}\, \xi_{2} \right)}-\frac{8 b w_{x 2}^{4}-w_{t2}^{2}+w_{x2}^{2}+w_{y2}^{2}}{2 w_{x2}^{2}},\label{(3.34a)}\\ 
		u(x,t) = \,&6 w_{x2}^{2} b \coth^{2}\left(\sqrt{-b}\, \xi_{2} \right)+\frac{6 w_{x2}^{2} b}{\coth^{2} \! \left(\sqrt{-b}\, \xi_{2} \right)}-\frac{8 b w_{x2}^{4}-w_{t 2}^{2}+w_{x2}^{2}+w_{y2}^{2}}{2 w_{x2}^{2}}, \label{(3.34b)} 
	\end{align}
\end{subequations}
where $\xi_{2}=t w_{t2}+y w_{y2}+x w_{x2}+b_{2}$.

If one chooses coefficients $\{b=-1, w_{x2} = 1, w_{y2} = 2,  w_{t2} = 1, b_{2}=1\}$ and take $t=1$ in the \Cref{(3.34a)}, we obtain \Cref{Figure_7}. Then the behavior of the solution can be intuitively expressed. \Cref{Figure_7}a is a three-dimensional diagram drawn based on the spatio-temporal domain $[-30, 30] \times [-30, 30]$. To observe the local  behavior of the solution, we plot the $x$-curves in \Cref{Figure_7}b at interval $t \in [-30, 30]$. And the contour and density plots of the solution are shown in \Cref{Figure_7}c and \Cref{Figure_7}d, respectively.

When $b>0$, 
\begin{subequations}\label{(3.35)}
	\begin{align}
		u(x,t) = &-6  w_{x2}^{2} b \tan^{2}\left(\sqrt{b}\, \xi_{2} \right)-\frac{6 w_{x2}^{2} b}{\tan^{2} \! \left(\sqrt{b}\, \xi_{2} \right)}-\frac{8 b w_{x2}^{4}-w_{t 2}^{2}+w_{x2}^{2}+w_{y2}^{2}}{2 w_{x2}^{2}}, \label{(3.35a)}\\
		u(x,t) = &-6 w_{x2}^{2} b \cot^{2}\left(\sqrt{b}\, \xi_{2}\right)-\frac{6 w_{x2}^{2} b}{\cot^{2} \! \left(\sqrt{b}\, \xi_{2} \right)}-\frac{8 b w_{x 2}^{4}-w_{t2}^{2}+w_{x2}^{2}+w_{y2}^{2}}{2 w_{x 2}^{2}}, 
	\end{align}
\end{subequations}
where $\xi_{2}=t w_{t2}+y w_{y2}+x w_{x2}+b_{2}$.
\begin{figure}[H]
	\centering
	\begin{subfigure}[b]{.32\linewidth} 
		\centering
		\includegraphics[scale=0.32]{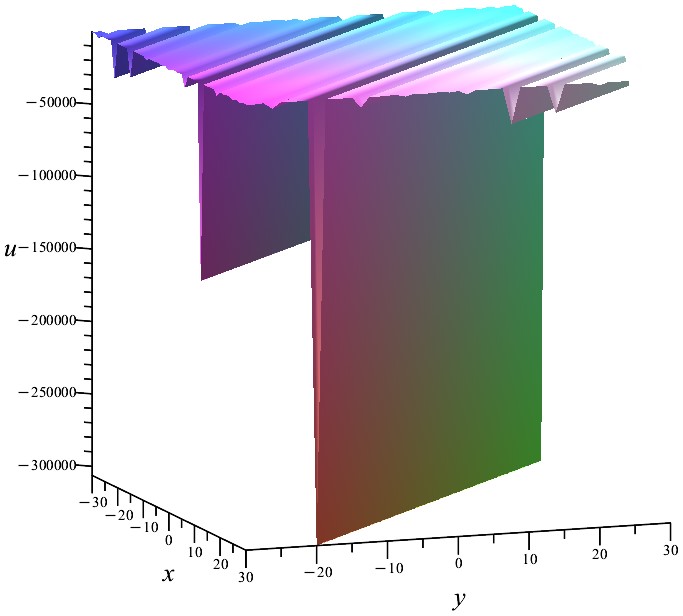}
		\subcaption{3D plot}
	\end{subfigure}
	\hfill 
	\begin{subfigure}[b]{.32\linewidth} 
		\centering
		\includegraphics[scale=0.33]{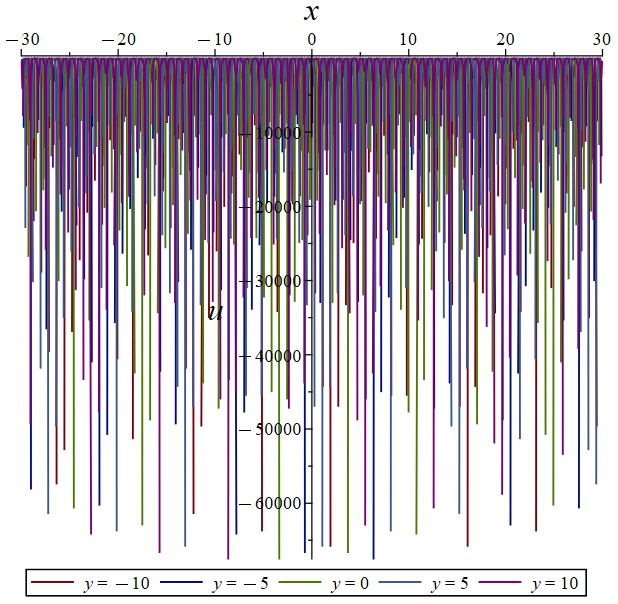}
		\subcaption{$x$-curves}
	\end{subfigure}
	\hfill 
	\begin{subfigure}[b]{.32\linewidth} 
		\centering
		\includegraphics[scale=0.33]{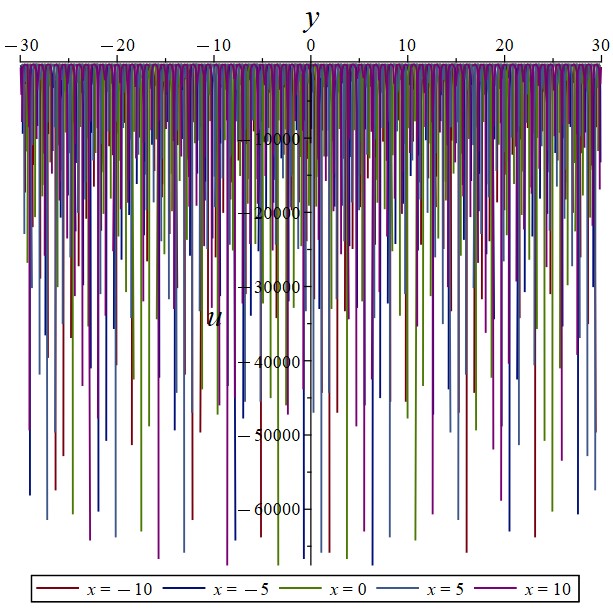}
		\subcaption{$y$-curves}
	\end{subfigure}
	
	\vspace{1em} 
	\begin{subfigure}[t]{.4\linewidth}
		\centering
		\includegraphics[scale=0.33]{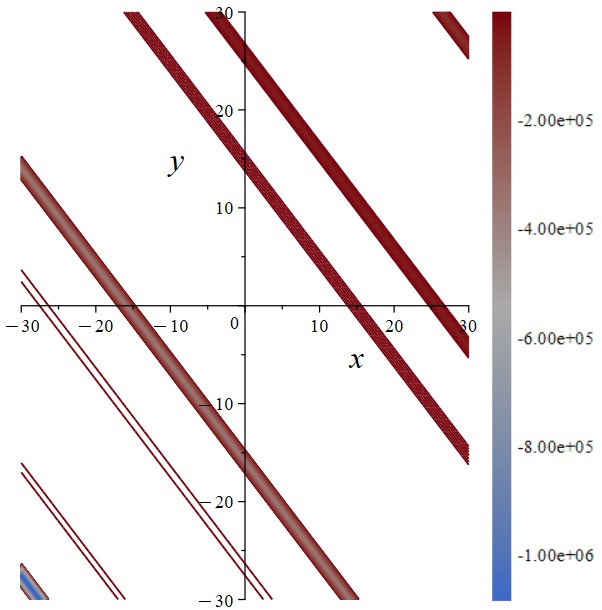}
		\subcaption{Contour plot}
	\end{subfigure}
	\hspace{-3em}
	\begin{subfigure}[t]{.4\linewidth}
		\centering
		\includegraphics[scale=0.33]{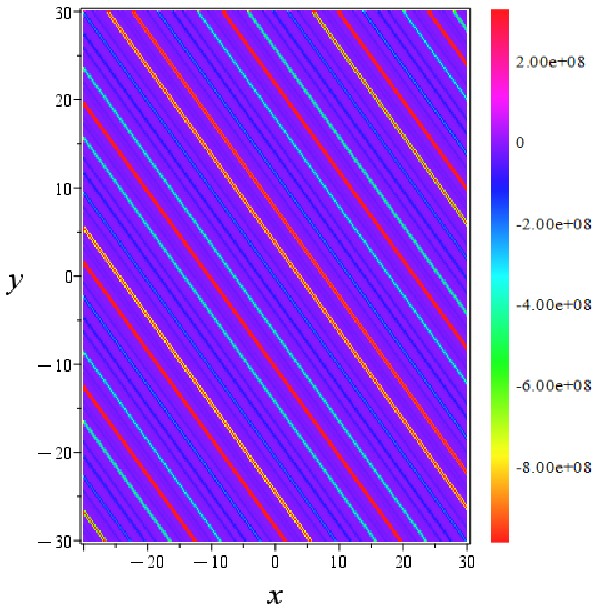}
		\subcaption{Density plot}
	\end{subfigure} 
	\centering
	\caption{The three-dimensional plot, curves plot, contour plot, and density plot of the solution \Cref{(3.35a)}.}
	\label{Figure_8}
\end{figure}

Similarly, to better observe the properties of the solution, we take the coefficients $\{b=1, w_{x2} = 2, w_{y2} = 2,  w_{t2} = 1, b_{2}=1\}$ and let $t=1$ in the \Cref{(3.35a)}. The three-dimensional plot, contour plot, and density plot of the solution are displayed in \Cref{Figure_8} based on the spatio-temporal domain $[-30, 30] \times [-30, 30]$.

\section{Comparison}\label{comparision}
In this section, nonlinear evolution \cref{(3.1)} are studied to demonstrate that the proposed method can not only obtain solutions already available in the literature but also yield novel and interesting results.

To further confirm the flexibility and customizability of our method, we employ different neural networks architectures to solve the \cref{(3.1)}. The 2-2-2-1 model is used to construct the potential analytical solution of equation, as shown in  \Cref{Figure_1}a.  Letting  $F_{1}(\xi_{1})=\varphi(\xi_{1}), F_{2}(\xi_{2})=\varphi(\xi_{2}), F_{3}(\xi_{3})=(\xi_{3}),$ and $F_{4}(\xi_{4})=(1/\xi_{4})$ in \Cref{Figure_1}a. And by substituting \Cref{(3.8)} into \Cref{(3.1)}, the following coefficient solutions are obtained via AENNM,

{\bf Solution 1:}
\begin{equation}\label{(4.1)}
	\left\{
	\begin{array}{l}
		b = b, 	
		\alpha = 
		\frac{-\gamma  w_{13}^{2} w_{3u}^{2}-2 w_{t1}^{2}}{2 w_{x1}^{2}}, 
		\beta = \gamma  b w_{13}^{2} w_{3u}^{2},
		b_{3} =  -\frac{b_{5}}{w_{3u}},
		b_{4} = 0, 
		b_{5} = b_{5},\\
		w_{13} = w_{13},
		w_{14} = w_{14},
		w_{23} = w_{24} = w_{4u} = 0,
		w_{3u} = w_{3u},
		w_{t1} = w_{t1},\\
		w_{t2} = w_{t2}, 
		w_{x1} = w_{x1}, 
		w_{x2} = w_{x2}.
	\end{array}
	\right\}
\end{equation}

By substituting \Cref{(4.1)} into \Cref{(3.8)}, the analytical solution of the original equation is obtained as follows,

\begin{equation}\label{(4.2)} 
		u(x,t)=w_{13} w_{3u}\, \varphi \! \left(t w_{t1}+x w_{x 1}+b_{1}\right). 
\end{equation}

When $b<0$, 
\begin{subequations}\label{(4.3)}
	\begin{align} 
		u(x,t)=&- w_{3u} w_{13} \sqrt{-b}\, \tanh \! \left(\sqrt{-b}\, \left(t w_{t1}+x w_{x1}+b_{1}\right)\right). \label{(4.3a)}\\
		u(x,t) =& -w_{3u} w_{13} \sqrt{-b}\, \coth \! \left(\sqrt{-b}\, \left(t w_{t1}+x w_{x 1}+b_{1}\right)\right). \label{(4.3b)}  
	\end{align}
\end{subequations}

When $b>0$, 
\begin{subequations}\label{(4.4)}
	\begin{align}
	    u(x,t) = \,&w_{3u} w_{13} \sqrt{b}\, \tan \! \left(\sqrt{b}\, \left(t w_{t1}+x w_{x1}+b_{1}\right)\right) \label{(4.4a)}.\\
		u(x,t) =\, &-w_{3u} w_{13} \sqrt{b}\, \cot \! \left(\sqrt{b}\, \left(t w_{t1}+x w_{x 1}+b_{1}\right)\right) \label{(4.4b)} .  
	\end{align}
\end{subequations}

If take the coefficients $\{b=-\beta/2(\alpha+\lambda^2), w_{t1} = \lambda,  b_{1}=0, w_{3u}w_{13}=\pm \sqrt{-2(\alpha+\lambda^2)/\gamma},\\ w_{x1} = 1 \}$  in the \Cref{(4.3a)} and \Cref{(4.4a)}, $\{b=-\beta/2(\alpha+\lambda^2), w_{t1} = \lambda,  w_{x1} = 1, w_{3u}w_{13}=\mp \sqrt{-2(\alpha+\lambda^2)/\gamma}, b_{1}=0\}$ in the \Cref{(4.3b)} and \Cref{(4.4b)}, the solutions obtained above are exactly the same as the true analytical solutions of \cite{elwakil2002modified} for \cref{(3.1)}. 

{\bf Solution 2:}
\begin{equation}\label{(4.5)}
	\left\{
	\begin{array}{l}
		b = b, 	
		\alpha = 
		-\frac{\gamma  w_{23}^{2} w_{3u}^{2}+2 w_{t2}^{2}}{2 w_{x 2}^{2}}, 
		\beta = -2 \gamma  b w_{23}^{2} w_{3u}^{2},
		b_{3} =  -\frac{b_{5}}{w_{3u}},
		b_{4} = 0, 
		b_{5} = b_{5},\\
		w_{13} = w_{14} = 0,
		w_{23} = w_{23},
		w_{24} = w_{24},
		w_{3u} = w_{3u},
		w_{4u} = b w_{23} w_{24} w_{3u},\\
		w_{t1} = w_{t1},
		w_{t2} = w_{t2}, 
		w_{x1} = w_{x1}, 
		w_{x2} = w_{x2}.
	\end{array}
	\right\}
\end{equation}

By substituting \cref{(4.5)} into \cref{(3.8)}, the solution of the original \cref{(3.1)} can be obtained as

\begin{equation}\label{(4.6)} 
	u(x,t)=w_{23}  w_{3u} \,\varphi \! \left(t w_{t2}+x w_{x 2}+b_{2}\right)+\frac{b w_{23} w_{3u}}{\varphi \! \left(t w_{t2}+x w_{x2}+b_{2}\right)}. 
\end{equation}

When $b<0$, 
\begin{equation}\label{(4.7)} 
	u(x,t)=\pm w_{23} w_{3u} \sqrt{-b}\, \mathrm{csch}\! \left(\sqrt{-b}\, \left(t w_{t2}+x w_{x 2}+b_{2}\right)\right) \mathrm{sech}\! \left(\sqrt{-b}\, \left(t w_{t2}+x w_{x 2}+b_{2}\right)\right).  
\end{equation}

When $b>0$, 
\begin{equation}\label{(4.8)} 
	u(x,t)=\pm w_{23} w_{3u} \sqrt{b}\,  \sec \! \left(\sqrt{b}\, \left(t w_{t2}+x w_{x2}+b_{2}\right)\right) \csc \! \left(\sqrt{b}\, \left(t w_{t2}+x w_{x 2}+b_{2}\right)\right).  
\end{equation}

If take the coefficients $\{b=\beta/4(\alpha+\lambda^2), w_{t2} = \lambda,  w_{x2} = 1, w_{23}w_{3u}= \sqrt{-2(\alpha+\lambda^2)/\gamma},\\ b_{2}=0\}$  in the \Cref{(4.7)}, $\{b=\beta/\alpha+\lambda^2, w_{t1} = \lambda,  w_{x1} = 1, w_{23}w_{3u}= \sqrt{-2(\alpha+\lambda^2)/\gamma}, b_{2}=0\}$ in the \Cref{(4.8)}, the solutions obtained above are exactly the same as the true analytical solutions of \cite{elwakil2002modified} for \cref{(3.1)}.

{\bf Solution 3:}
\begin{equation}\label{(4.9)}
	\left\{
	\begin{array}{l}
		b = b, 	
		\alpha = 
		-\frac{\gamma  w_{23}^{2} w_{3u}^{2}+2 w_{t2}^{2}}{2 w_{x 2}^{2}}, 
		\beta = 4 \gamma  b w_{23}^{2} w_{3u}^{2},
		b_{3} =  -\frac{b_{5}}{w_{3u}},
		b_{4} = 0, 
		b_{5} = b_{5},\\
		w_{13} = w_{14} = 0,
		w_{23} = w_{23},
		w_{24} = w_{24},
		w_{3u} = w_{3u},
		w_{4u} = -b w_{23} w_{24} w_{3u},\\
		w_{t1} = w_{t1},
		w_{t2} = w_{t2}, 
		w_{x1} = w_{x1}, 
		w_{x2} = w_{x2}.
	\end{array}
	\right\}
\end{equation}

By substituting \Cref{(4.9)} into \Cref{(3.8)}, the analytical solution of the original equation is obtained as follows,

\begin{equation}\label{(4.10)} 
	u(x,t) = w_{23} w_{3u} \varphi \! \left(t w_{t2}+x w_{x2}+b_{2}\right) -\frac{b w_{23} w_{3u}}{\varphi \! \left(t w_{t2}+x w_{x 2}+b_{2}\right)}
	. 
\end{equation}

When $b<0$, 
\begin{equation}\label{(4.11)}
		u(x,t)=-w_{23} w_{3u} \sqrt{-b} \left(\tanh \! \left(\sqrt{-b}\, \xi_{2}\right)+\coth \! \left(\sqrt{-b}\, \xi_{2}\right)\right),
\end{equation}
where $\xi_{2}=t w_{t2}+x w_{x2}+b_{2}$.

\vspace{3pt}
When $b>0$, 
\begin{equation}\label{(4.12)}
	u(x,t)=w_{23} w_{3u} \sqrt{b}\,  \left(\tan \! \left(\sqrt{b}\, \xi_{2} \right)-\cot \! \left(\sqrt{b}\, \xi_{2} \right)\right),  
\end{equation}
where $\xi_{2}=t w_{t2}+x w_{x2}+b_{2}$.

If take the coefficients $\{b=-\beta/8(\alpha+\lambda^2), w_{t2} = \lambda,  b_{2}=0, w_{23}w_{3u}=\mp \sqrt{-2(\alpha+\lambda^2)/\gamma},\\w_{x2} = 1 \}$  in the \Cref{(4.11)} and \Cref{(4.12)}, the solutions obtained above are exactly the same as the true analytical solutions of \cite{elwakil2002modified} for \cref{(3.1)}.

Therefore, the solution obtained by the proposed method can yield the solutions derived using the modified extended tanh-function method \cite{elwakil2002modified} by taking specific values for the coefficients. Moreover, the solutions obtained by AENNM not only include those derived from existing methods in the literature but can also yield new solutions, such as \cref{(3.5)} and \cref{(3.6)}.

\vspace{3pt}
\noindent{\bf Remark 1.} This paper presents a variety of NNs architectures designed for obtaining analytical solutions to NLPDEs. Owing to the method's significant flexibility and adaptability, different networks configurations can be employed to derive novel types of solutions for these equations.

\noindent{\bf Remark 2.} To verify the reliability of the results, the derived analytical solutions are substituted back into the original equations, confirming that both sides remain equivalent. All solutions obtained by the analytical method presented in this study have been rigorously validated using Maple software.

\section{Discussions}\label{discussions} 
This paper proposes a novel method that combines NNs models with symbolic computation techniques to rapidly obtain exact analytical solutions for NLPDEs. Compared to traditional NNs approaches \cite{Hornik1989Multilayer,raissi2019physics,lu2021operators,wang2021operators}, this method solely utilizes the feedforward computation of NNs to construct trial functions for the equations, without incorporating the training mechanism of NNs. Therefore, the proposed approach eliminates the need for iterative calculations and optimization procedures. Furthermore, this approach yields exact solutions to the equations without data samples, ensuring zero approximation error. This paper introduces a novel approach to obtain exact solutions of NLPDEs which has the potential to advance the field by providing solutions without the need for large datasets or iterative optimization.

Traditional symbolic computation methods, such as the extended tanh-function metho-d, are applicable to NLPDEs that can be transformed into ordinary differential equations via traveling wave transformations. When using such methods, the form of the preset solution is fixed. Moreover, traditional symbolic computation approaches require tailored solution strategies for different equations. In contrast, the proposed method in this paper does not rely on traveling wave transformations but instead directly employs a NNs structure to construct trial functions for the given equations. By leveraging the NNs architecture, the design of trial functions becomes more systematic and explicit. 

Recently, the bilinear neural network method (BNNM) \cite{Zhang2019Bilinear} has been successfully applied to solve various NLPDEs, including the Caudrey-Dodd-Gibbon-Kotera-Sawada-like equation \cite{Zhang2021Generalized}, the Boussinesq  equation \cite{Isah2024Exploring}, the extended Sawada-Kotera equation \cite{Gai2024N-solitons}, and the Boiti–Leon–Manna–Pempinelli-like equation \cite{Qin2024Various}. However, BNNM requires complex preprocessing of the equations, namely bilinear transformation, which poses a challenge for researchers without a strong mathematical background. Moreover, not all partial differential equations possess a bilinear form, further limiting the applicability of this method. Unlike BNNM, the method proposed in this paper does not require bilinear transformation of the equations, thereby lowering the barrier for researchers.

However, the proposed method also has certain limitations as follows: (i) Designing appropriate neural networks architectures is crucial for solving NLPDEs. Only by establishing suitable networks structures can optimal solutions be obtained. (ii) As the complexity of neural networks models increases, the computational workload grows correspondingly.
\section{Conclusions}\label{conclusions} 
In this study, we propose the auxiliary equation neural networks method (AENNM) to analytically solve NLPDEs. This method introduces a novel activation function derived from solutions of the Riccati equation for NNs, establishing a new connection between differential equations and deep learning. The method introduces a novel interpretable activation function for NNs, enabling the systematic construction of new trial functions for NLPDEs within the "2-2-2-1" and "3-2-2-1" networks architectures. The feasibility of this method is demonstrated by solving the nonlinear evolution equation, the Korteweg-de Vries-Burgers equation, and the (2+1)-dimensional Boussinesq equation. Due to embedding the auxiliary equation method into the NNs framework, some new and interesting solutions are obtained. The solutions expressed in terms of hyperbolic, trigonometric, and rational functions provide valuable insights into the dynamic behaviors of the systems described by these equations. Three-dimensional plots, contour plots, and density plots are given to observe the dynamic behaviors of the obtained solutions.

Compared to traditional neural network methods, AENNM effectively balances the strong expressive power of neural networks with the high precision of symbolic computation. As a result, the proposed method achieves enhanced computational efficiency while maintaining high accuracy. Furthermore, the flexibility and adaptability of AENNM allow it to be applied to a wide range of NLPDEs by adjusting the network architecture, including the number of layers, neurons, and activation functions. The proposed method does not rely on traveling wave transformations, but instead embeds the solution of the equation as an activation function within the neural network architecture. Compared to exsiting methods in the literature, AENNM not only recovers known solutions but also discovers previously unreported ones, thereby enhancing our understanding of NLPDEs.
The successful application of AENNM highlights its potential as a powerful tool for tackling complex nonlinear problems across diverse fields such as physics, engineering, and applied mathematics. Future research may focus on further refining the method, exploring its applicability to more intricate equations, and potentially integrating it with other advanced techniques to expand its capabilities.

\vspace{8pt}
\noindent{\bf Acknowledgements}

\vspace{8pt}
 This work was supported  by the Natural Science Foundation of Shandong Province under Grant No.ZR2023MA062, the Belt and Road Special Foundation of The National Key Laboratory of Water Disaster Prevention under Grant No.2023491911, Tianyuan Fund for Mathematics of the National Natural Science Foundation of China under Grant No.12426105, and the Scientific and Technological Innovation Programs (STIP) of Higher Education Institutions in Shanxi under Grant No.2024L022.

\vspace{8pt}
\noindent{\bf Data availability}

\vspace{8pt}
No data was used for the research described in the article.

\vspace{8pt}
\noindent{\bf Conflict of interest}

\vspace{8pt}
The authors declare that they have no conflicts of interest.

\end{document}